\documentclass[letterpaper, 10 pt, conference]{ieeeconf}  
\IEEEoverridecommandlockouts                              

\let\labelindent\relax
\usepackage{amsmath, amsthm, amssymb}
\usepackage{xcolor}
\usepackage{todonotes}
\usepackage{graphicx}
\usepackage{hyperref}
\usepackage[inline]{enumitem}
\usepackage[space, compress, sort]{cite}
\usepackage{multirow}
\usepackage{xspace}
\usepackage{booktabs}
\newcommand{\Method}{DroneFL\xspace}

\usepackage[T1]{fontenc}

\title{\LARGE \bf \Method: Federated Learning for Multi-UAV Visual Target Tracking}

\author{Xiaofan Yu$^{1,3}$, Yuwei Wu$^{2}$, Katherine Mao$^{2}$, Ye Tian$^{1}$, Vijay Kumar$^{2}$, Tajana Rosing$^{1}$
\thanks{*This research was supported in part by the National Science Foundation under Grants \#2112665 (TILOS AI Research Institute), \#2003279, \#1826967, \#2100237, \#2112167, \#1911095, \#2120019, \#2211386 and in part by PRISM and CoCoSys, centers in JUMP 2.0, an SRC program sponsored by DARPA.}
\thanks{$^{1}$X. Yu, Y. Tian and T. Rosing are with the Department of Computer Science and Engineeering, University of California at San Diego, La Jolla, CA 92093, USA. E-mail: \texttt{\{x1yu, yet002, tajana\}@ucsd.edu}.}
\thanks{$^{2}$Y. Wu, K. Mao and V. Kumar are with the GRASP Lab, University of Pennsylvania, Philadelphia, PA 19104, USA. E-mail: \texttt{\{yuweiwu, maokat, kumar\}@seas.upenn.edu}.}
\thanks{$^{3}$X. Yu is also with the Department of Electrical Engineering, University of California at Merced, Merced, CA 95343, USA.}
}

\begin{document}

\maketitle
\thispagestyle{empty}
\pagestyle{empty}

\begin{abstract}
Multi-robot target tracking is a fundamental problem that requires coordinated monitoring of dynamic entities in applications such as precision agriculture, environmental monitoring, disaster response, and security surveillance.
While Federated Learning (FL) has the potential to enhance learning across multiple robots without centralized data aggregation, its use in multi-Unmanned Aerial Vehicle (UAV) target tracking remains largely underexplored.
Key challenges include limited onboard computational resources, significant data heterogeneity in FL due to varying targets and the fields of view, and the need for tight coupling between trajectory prediction and multi-robot planning.
In this paper, we introduce \Method, the first federated learning framework specifically designed for efficient multi-UAV target tracking.
We design a lightweight local model to predict target trajectories from sensor inputs, using a frozen YOLO backbone and a shallow transformer for efficient onboard training.
The updated models are periodically aggregated in the cloud for global knowledge sharing.
To alleviate the data heterogeneity that hinders FL convergence, \Method introduces a position-invariant model architecture with altitude-based adaptive instance normalization.
Finally, we fuse predictions from multiple UAVs in the cloud and generate optimal trajectories that balance target prediction accuracy and overall tracking performance.
Our results show that \Method reduces prediction error by 6\%-83\% and tracking distance by 0.4\%-4.6\% compared to a distributed non-FL framework. In terms of efficiency, \Method runs in real time on a Raspberry Pi 5 and has on average just 1.56 KBps data rate to the cloud.

\end{abstract}

\section{Introduction}
\label{sec:intro}

Multi-robot systems are increasingly deployed in real-world applications, with aerial platforms playing a distinct role in large-scale environments such as industry, agriculture, and environmental monitoring~\cite{ https://doi.org/10.1002/rob.20403, 9476820}.
Teams of UAVs provide wide-area coverage, maintaining visibility even in cluttered, GPS-denied settings like orchards, while adapting quickly to changing operational demands. 
In agriculture, for example, UAVs enable large-scale monitoring by tracking ground vehicles during operations such as harvesting. 
Unlike static sensors, these targets have dynamic motions influenced by heterogeneous machinery, variable field conditions, and operator-dependent behaviors. 
Therefore, accurate monitoring requires coordinated efforts across multiple UAVs.

\begin{figure}
    \centering
    \vspace{2mm}
    \includegraphics[width=1.0\linewidth]{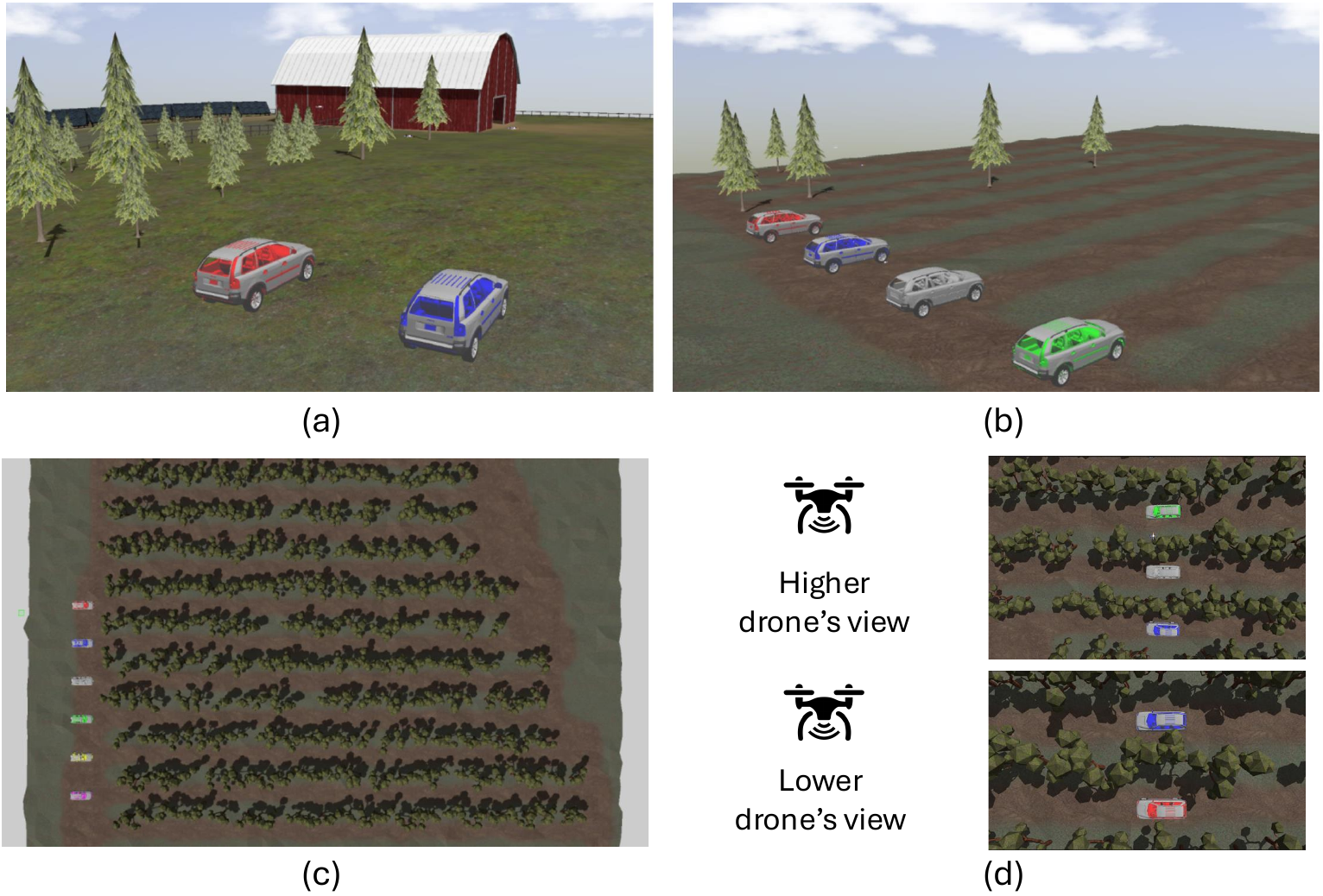}
    \vspace{-8mm}
    \caption{Simulated multi-robot target tracking testbed using agricultural setting as a case study. (a–c) Different environment setups from multiple views. Distinct targets are indicated by colors. In the orchard environment (c), the vehicles move through the rows while performing agricultural operations, and they are treated as tracked targets. (d) Example RGB images from drone-mounted cameras at different flight altitudes.}
    \label{fig:teaser}
    \vspace{-6mm}
\end{figure}

In this paper, we consider multi-robot target tracking as a representative scenario (Fig.~\ref{fig:teaser}).
Most existing research~\cite{ chen2025distributed,corah2021scalable,zhou2022graph,liu2022decentralized,li2024resilient} relies on low-dimensional sensors and fixed models or heuristics, which fail to handle high-dimensional inputs such as images and video streams, or to adapt to the variability of real-world environments.
In addition, UAVs face strict Size, Weight, and Power (SWaP) constraints that limit both onboard computation and communication. These challenges motivate the need for a multi-robot framework that processes rich visual inputs, delivers high performance (e.g., prediction accuracy and tracking quality) across varying environments, while minimizing resource consumption.

Federated learning (FL)~\cite{mcmahan2017communication} provides a promising solution by allowing distributed UAVs to collaboratively learn a vision-based neural network (NN) model.
Each FL round begins with a central server (cloud) distributing a global model to the robots. The robots then update their NNs using locally collected sensor data to better predict target trajectories. Finally, the updated models are sent back to the cloud for aggregation, enabling generalization across multiple robots.
During inference, each robot sends its target trajectory predictions to the central server. The server fuses these predictions to form a global understanding of target movements and generate future trajectories for the UAV team.
Compared to fixed models or heuristics, FL allows on-device adaptation to environmental variations, leading to better long-term performance. Unlike centralized approaches that send all raw data to the cloud, FL supports real-time inference and reduces communication overhead by keeping high-dimensional data local.
Although deploying and training NN models on edge devices was a major challenge in the past, recent advances in efficient algorithms and more capable hardware have made it increasingly feasible.
For instance, the YOLOv11n model can process $640\times640$ images on a Raspberry Pi 5 in around 100 ms per image~\cite{yolo-rpi5}.
Overall, an FL-enabled multi-robot target tracking framework offers the potential to improve scalability, robustness, and communication efficiency compared to traditional approaches.

While FL is promising, several key challenges remain in designing a practical FL framework for multi-UAV target tracking. These challenges include:
\begin{enumerate}[label=(\arabic*),noitemsep,nolistsep]
  \item \textbf{Limited device resources:} Despite recent advances, deploying NNs on UAVs remains challenging due to their high resource demands for training and inference~\cite{wang2019deep}. Wireless communication with the cloud also consumes significant energy~\cite{luo2021cost}. 
  The FL framework should minimize resource consumption and extend the endurance of SWaP-constrained UAVs.
  
  \item \textbf{Data heterogeneity in FL:} Data heterogeneity, namely the varying distributions of data collected by distributed devices, is a long-standing challenge in FL~\cite{li2020fedprox,wang2020tackling,karimireddy2020scaffold}. 
  In visual-based target tracking, data heterogeneity may result from variations in monitored targets or differences in the robots’ fields of view.
  If not properly addressed, data heterogeneity can hinder learning convergence in FL and degrade trajectory prediction performance.
  
  \item \textbf{Robust prediction-action coupling:} 
  Multi-UAV tracking requires integrating trajectory prediction with trajectory planning: accurate predictions are critical for generating effective maneuvers, while the chosen actions in turn influence future observations, making this closed-loop coupling essential for long-term performance.
  
\end{enumerate}
Existing studies on FL for distributed trajectory prediction~\cite{majcherczyk2021flow,peng2023privacy,wang2024federated,tao2024scenario,benhelal2024siamfltp} have primarily focused on autonomous driving, where data heterogeneity is the main challenge. 
In contrast, a complete framework that applies FL to multi-UAV target tracking, while addressing SWaP constraints, data heterogeneity and real-time control, is still lacking.

In this paper, we propose \Method, an end-to-end federated learning framework for multi-robot target tracking using vision-based perception. \Method addresses the above challenges through three key designs spanning target trajectory prediction and drone trajectory planning. First, we introduce a lightweight neural network architecture to operate within SWaP constraints. Each robot uses a frozen YOLO model~\cite{yolo-rpi5} to extract bounding box detections from images. The bounding boxes and recent odometry data are then fed into a shallow transformer model~\cite{vaswani2017attention} that predicts future target trajectories. During FL, only the transformer layers are updated on the drone, minimizing onboard latency and energy consumption. Onboard storage is also reduced by storing only bounding boxes instead of raw image data. Second, we observe that FL convergence is mainly hindered by data heterogeneity caused by flight altitudes, which produce different field-of-view patterns. To address this, we propose an altitude-based adaptive instance normalization design in the transformer model, making it position-invariant and easier to train across drones with varying views.
Finally, we develop a centralized trajectory planning module that fuses predictions from multiple robots using an Extended Kalman Filter with adaptive innovation, and generates optimal trajectories by minimizing a custom cost function that balances tracking robustness with perception quality.
\Method is extensively evaluated in a simulated testbed, using an agricultural setting as a case study. \Method is further validated for efficiency on a Raspberry Pi 5, a representative computing unit for resource-constrained UAVs.

In summary, the contributions of this paper are:
\begin{itemize}
\item \Method, the first FL framework for multi-UAV target tracking, which tightly couples closed-loop trajectory prediction and trajectory planning through efficient and effective FL under data heterogeneity

\item Comprehensive evaluations showing that \Method reduces prediction error by 6\%-83\% and tracking distance by 0.4\%-4.6 \% compared to distributed non-FL framework. \Method achieves real-time prediction on a representative computing unit for UAVs, with only $1.56$ KB/s communication with the cloud.
\end{itemize}

\section{Related Work}
\label{sec:related-work}

\subsection{Federated Learning}

In recent years, FL has emerged as a popular framework for distributed learning
The standard FL algorithm, FedAvg~\cite{mcmahan2017communication}, aggregates client models by averaging their weights.
Existing works have enhanced FedAvg to tackle various heterogeneity challenges, including non-iid data distribution~\cite{li2020fedprox,karimireddy2020scaffold}, diverse system resources~\cite{li2022pyramidfl}, heterogeneous sensor modalities~\cite{zhao2024multimodalhd}, and varying model architectures~\cite{zhang2024few}.

Prior FL research for multi-robot applications has focused on predicting future trajectories from past ones in autonomous driving.
Flow-FL~\cite{majcherczyk2021flow} introduced the first open-source FL dataset for swarm motion with communication graph. ALFLTP~\cite{peng2023privacy} developed the first FL framework for vehicle trajectory prediction, integrating an aleatoric uncertainty metric for active client selection. More recently, studies have explored FL on roadside units or drones for vehicle trajectory predictions~\cite{wang2024federated,tao2024scenario,benhelal2024siamfltp}.
While these works have advanced FL for multi-robot trajectory prediction, they do not process raw images or incorporate robot control. 
In contrast, \Method closes the loop by integrating vision-based target trajectory prediction with UAV trajectory planning.

\subsection{Visual-based Multi-Robot Target Tracking}

Incorporating visual constraints into planning has become increasingly important in aerial robotics applications.
Visual control and perception-aware planning are widely used in tasks like inspection and target tracking~\cite{7921549, 9197157}. 
When extended to multi-robot problems, the complexity increases due to challenges such as inter-robot task allocation, decentralized sensing, and shared exposure to environmental hazards.
The work in~\cite{chen2025distributed} integrates diverse sensors within a cooperative framework, using metrics to balance effectiveness across modalities.
Corah \textit{et al.}~\cite{corah2021scalable} improves scalability with a decentralized planner that approximates centralized performance using local computations. 
In addition,~\cite{liu2022decentralized, li2024resilient, mayya2022adaptive}
address the environmental dangers by exploring trade-offs that balance maximizing team performance with minimizing the risk of sensor degradation or failure.
While these works improve robustness, scalability, and sensing efficiency, most existing systems rely on locally trained models or heuristics, limiting their generalizability in real-world environments.
In contrast, \Method incorporates FL on the UAV to improve generalization ability.

\section{Problem Formulation}
\label{sec:problem}

\begin{figure}
    \centering
    \vspace{1mm}
    \includegraphics[width=0.9\linewidth]{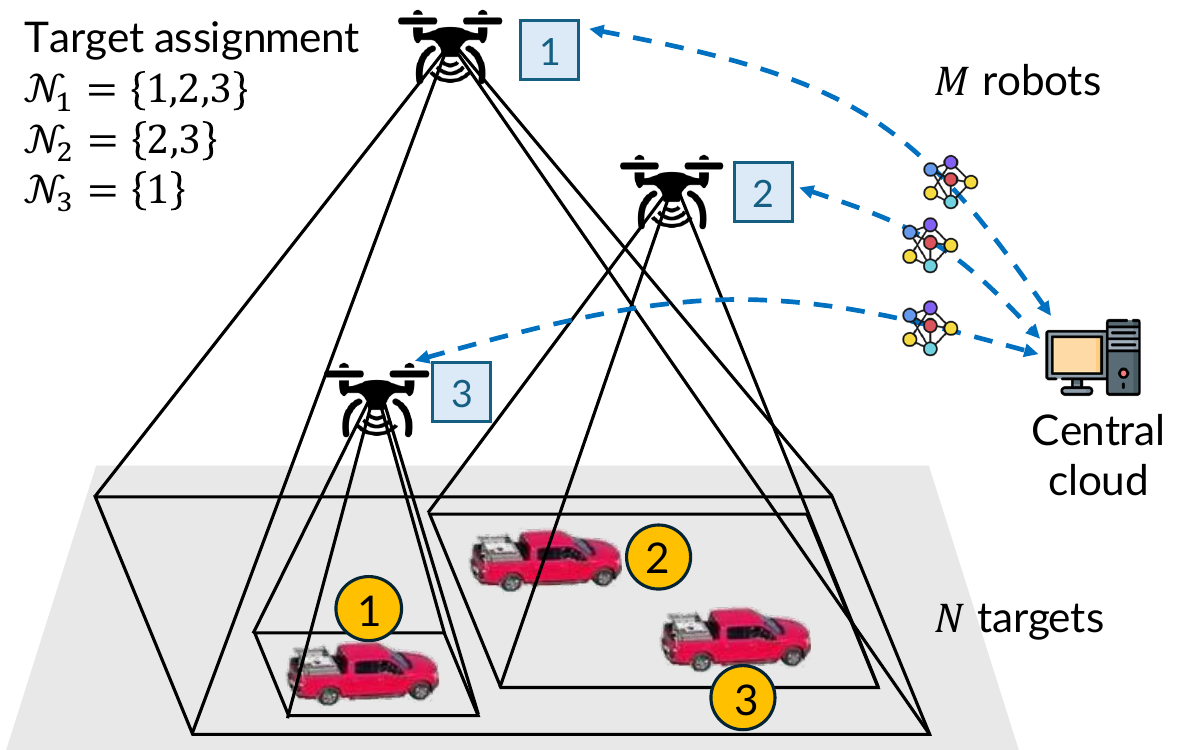}
    \vspace{-3mm}
    \caption{Example problem setup with $M=3$ robots and $N=3$ targets. Each robot $i$ is assigned a subset of targets $\mathcal{N}_i$ to track and flies at a distinct fixed altitude to avoid conflicts. All robots communicate wirelessly with a central cloud for FL, as indicated by the blue dashed line.}
    \label{fig:problem_diagram}
    \vspace{-4mm}
\end{figure}

We consider a team of $M$ aerial robots tasked with monitoring $N$ targets, as illustrated in Fig.~\ref{fig:problem_diagram}. 
Each robot is equipped with onboard sensors for target detection. 
The objective of the team is to ensure reliable tracking performance. For example, minimizing prediction error, reducing trajectory uncertainty, and/or maximizing target coverage.
Formally, we define the coupled problem as:
\begin{equation}
\min_{\phi, \mathbf{u}_{1:T}} \; \mathbb{E}\!\left[ \mathcal{J}(\hat{\mathbf{z}}_{1:T}(\phi), \mathbf{z}_{1:T}, \mathbf{u}_{1:T}) \right],
\label{eq:main-objective}
\end{equation}
where $\phi$ is the predictive model, $\hat{\mathbf{z}}_{1:T}(\phi)$ are the predicted trajectories, $\mathbf{z}_{1:T}$ are the reference trajectories, $\mathbf{u}_{1:T}$ are the robot control inputs, and $\mathcal{J}$ is a task-specific cost (e.g., error, uncertainty, or coverage).
To achieve the goal, our framework addresses two coupled subproblems: (1) learning a predictive model $\phi$ to forecast future target positions, and (2) computing the control inputs $\mathbf{u}_t$ for all robots based on these predictions. 
This objective consists of two coupled subproblems: (1) learning a predictive model $\phi$ via Federated Learning (Sec.~\ref{sec:problem-fl}), and (2) computing control inputs $\mathbf{u}_t$ for all robots based on predicted trajectories (Sec.~\ref{sec:problem-mrtt}).

\subsection{Federated Learning for Trajectory Prediction}
\label{sec:problem-fl}

In the FL setup, each robot maintains an onboard dataset from past observations, with $\mathcal{T}_i$ denoting the set of valid data timestamps on robot $i$.
Formally, let $\mathbf{x}^i_{t}$ denote the odometry of robot $i \in [M]$ at time $t \in \mathcal{T}_i$, and let $\gamma^i_{t}$ be the RGB image captured by its camera. We use $[n]$ to denote a set of consecutive integers from $1$ to $n$. 
From odometry and visual inputs, the neural network model $\phi$ predicts the position of target $j \in [N]$, denoted $\hat{\mathbf{z}}^i_{j,t}$. 
For supervised training in simulation, reference trajectories $\mathbf{z}_{j,t}$ are available from the simulator or external localization.
During online deployments, where direct ground truth is unavailable, robots rely on historical estimates or pseudo-labels generated from past tracking results to update local models. The cloud is a central desktop located within the communication range of all UAVs for model aggregation.

For any $t \in \mathcal{T}_i$, define the past time window of length $L_p$ as $t_p = \{t-L_p+1,\ldots,t\}$ and the future window of length $L_f$ as $t_f = \{t+1,\ldots,t+L_f\}$. $\mathcal{N}_i$ denotes the set of targets assigned to robot $i$. The first trajectory prediction task is formulated as:
\begin{subequations}
\label{eq:fl-problem}
\begin{align}
\min_\phi \quad \mathcal{J}_{\text{pred}}(\phi) &= \frac{1}{M} \sum_{i \in [M]} \frac{1}{|\mathcal{N}_i| |\mathcal{T}_i|} \sum_{j \in {\mathcal{N}_i}} \sum_{t \in \mathcal{T}_i} \left\| \mathbf{\hat{z}}^i_{j,t_f} - \mathbf{z}_{j,t_f} \right\| \label{eq:fl-obj} \\
\textrm{s.t.} \quad \mathbf{\hat{z}}^i_{j,t_f} &= \phi ( \mathbf{x}^i_{t_p}, \gamma^i_{t_p}) \quad \forall t \in \mathcal{T}_i \label{eq:perception}
\end{align}
\end{subequations}
Equation~\eqref{eq:fl-obj} defines the federated learning objective of minimizing each robot’s prediction error on its assigned targets using past data. Equation~\eqref{eq:perception} describes the trajectory prediction process on robot $i$ using model $\phi$. This problem can be addressed through federated learning: each robot minimizes the prediction error for its assigned targets, while local models are periodically shared and aggregated in the cloud to combine knowledge from all robots.

\subsection{Multi-Robot Target Tracking}
\label{sec:problem-mrtt}

Given predicted positions $\hat{\mathbf{z}}_{t_f}$, robot trajectories are determined by optimizing over a planning horizon. 
Let $\mathbf{p}_t$ denote robot positions at time $t$, and $\mathbf{u}_{t_f} = [\mathbf{u}^1_{t_f}, \ldots, \mathbf{u}^M_{t_f}]^\intercal$ the control inputs for all robots. The tracking problem is defined as an optimization problem:
\begin{subequations}
\label{eq:control}
\begin{align}
    \min_{\mathbf{u}_{t_f}} \quad & \mathcal{J}_{\text{track}}(\mathbf{p}_{t_f}, \hat{\mathbf{z}}_{t_f}) =  Q(\mathbf{p}_{t_f}, \hat{\mathbf{z}}_{t_f}) + \omega \sum_{i \in [M]} \left\| \mathbf{u}^i_{t_f} \right\| \label{eq:obj} \\
    \textrm{s.t.} \quad & \mathbf{p}_{t+1} = \mathbf{\Phi} \mathbf{p}_t + \mathbf{\Lambda} \mathbf{u}_t \quad \forall t \in t_f \label{eq:measurement} \\
     \qquad & \mathbf{u}_{t_f} \in \mathcal{U}, \; \mathbf{p}_t \in \mathcal{P}
\end{align}
\end{subequations}
\noindent where $Q$ measures tracking performance (e.g., error or uncertainty), and the second term penalizes control effort, with a weight $\omega$ to balance the tradeoff.
The dynamics constraint in~\eqref{eq:measurement} models robot motion, while $\mathcal{U}$ and $\mathcal{P}$ denote feasible control and position sets.
Our formulation also allows possible extensions such as additional constraints for area restrictions and danger zones~\cite{liu2022decentralized,li2024resilient}.

An effective solution requires coordination between the two subproblems. To this end, we present \Method, an end-to-end framework that unifies federated trajectory prediction with multi-robot target tracking.

\section{System Framework}
\label{sec:method}

\begin{figure*}[t]
    \centering
    \includegraphics[width=0.97\linewidth]{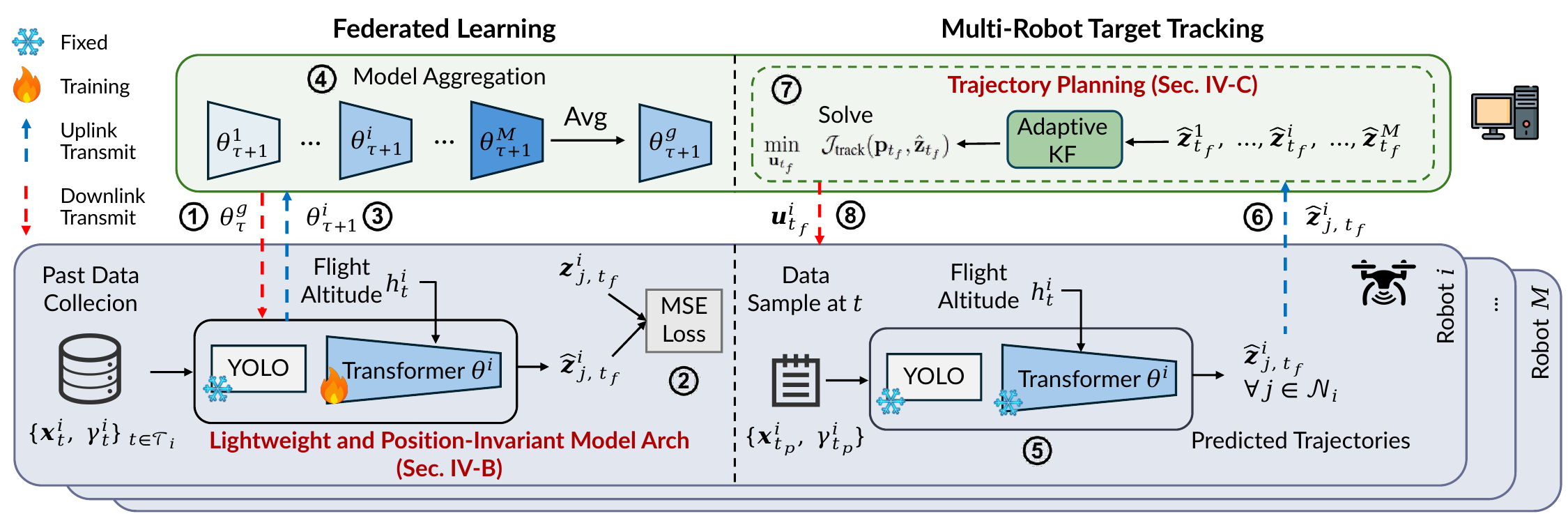}
    \vspace{-3mm}
    \caption{Overall diagram of \Method showing the central desktop (top) and deployments on robots (bottom). \Method consists of two main components: federated learning (left) and multi-robot target tracking (right), both relying on uplink and downlink communication between the desktop and robots. The two key modules, highlighted in red, are detailed in Sec.~\ref{sec:method-model-arch} and \ref{sec:method-control}.} 
    \label{fig:method}
    \vspace{-4mm}
\end{figure*}

In this section, we present the system framework and detailed design of \Method.
\Method addresses three key challenges: (1) ensuring computational and communication efficiency with a lightweight trajectory prediction model for FL, (2) handling data heterogeneity through a position-invariant architecture, and (3) enabling coordination via adaptive prediction fusion and trajectory optimization to balance target prediction and tracking.

\subsection{Overview of \Method}

\label{sec:method-overview}
Fig.~\ref{fig:method} shows the overall architecture of \Method.
As a distributed system, \Method spans a central desktop (cloud) and a team of aerial robots connected via wireless communication.
\Method has two main components: federated learning and trajectory planning, which run in parallel spanning both robots and the cloud.
The FL component trains a target trajectory prediction model $\phi$ on each robot’s onboard dataset without large quantities of raw data. It runs periodically or when onboard errors increase, but not in real time. In contrast, the trajectory planning component computes future UAV trajectories in real time. The scheduling of these two components balances real-time tracking performance and resource use. We next detail the step-by-step execution of each component.

In the FL process, each global round begins with the central cloud sending the latest global model $\theta^g_\tau$ to all robots (\textcircled{1} in Fig.~\ref{fig:method}), where $\tau$ is the round index. As described in Sec.~\ref{sec:method-model-arch}, the model architecture $\phi$ combines a frozen YOLO backbone~\cite{yolo-rpi5} with a trainable shallow transformer $\theta$~\cite{vaswani2017attention}. Thus, only the trainable weights are exchanged with the cloud. After receiving the global weights, each robot updates its local model and trains it for $E$ epochs on past onboard data. The training loss is the MSE between predicted and ground-truth trajectories (\textcircled{2}). Once training is complete, the updated weights $\theta^i_{\tau+1}$ are sent back to the cloud.
Unlike traditional FL, which selects a subset of clients for local training~\cite{mcmahan2017communication,li2020fedprox}, \Method involves all connected robots, as the system operates with only a small number of robots rather than hundreds of clients. After collecting updates from every robot, the cloud performs a canonical FedAvg aggregation~\cite{mcmahan2017communication} by averaging the models to obtain the new global model $\theta^g_{\tau+1}$ (\textcircled{4}).
We emphasize that \Method focuses on addressing efficiency and data heterogeneity in FL for multi-robot target tracking, rather than client selection or model aggregation. These aspects are well-studied in prior work~\cite{cho2022towards,fu2023client,yu2023async,pillutla2022robust,ma2022layer}, and their techniques can be seamlessly integrated with \Method in future extensions.

For real-time trajectory planning, \Method designs a centralized planner for its simplicity and improved coordination among robots. 
The process starts when a new sample arrives. The neural network model predicts the trajectory in the future time window $t_f$ (\textcircled{5}). Each robot then transmits its predictions of the targets in view to the central cloud (\textcircled{6}). The core design of trajectory planning in \Method is a centralized planner that uses an Extended Kalman Filter to estimate prediction noise (\textcircled{7}), compute the cost function $Q(\mathbf{p}_{t_f}, \hat{\mathbf{z}}_{t_f})$ and select the optimal control input. Finally, the control inputs are sent back to each robot for real-time execution (\textcircled{8}). 
Note that \Method minimizes the communication costs by only transmitting the lightweight transformer model, the predicted positions, and the control decisions, while avoiding communicating raw data. We leave the design of a distributed control policy to future work. 

We next provide the details of two novel \Method contributions: the lightweight and position-invariant model architecture (Sec.~\ref{sec:method-model-arch}) and the centralized planner for multi-robot target tracking (Sec.~\ref{sec:method-control}).

\subsection{Lightweight and Position-Invariant Model Architecture}
\label{sec:method-model-arch}

The FL design in \Method faces two main challenges: limited onboard resources and data heterogeneity across robots. To overcome them, \Method introduces a lightweight model architecture and a position-invariant design.

First, we design the trajectory prediction model $\phi$ using a frozen YOLO detector~\cite{yolo-rpi5} and a shallow transformer architecture $\theta$~\cite{vaswani2017attention}, as shown in Fig.~\ref{fig:method}. 
This design avoids the heavy resource consumption of training a complex image processing network onboard. The frozen YOLO detector is pretrained offline for accurate target detection, and each detected bounding box is assigned a target label. The shallow transformer predictor then takes the bounding boxes from YOLO, combined with the odometry (including drone positions), and predicts the future positions of the corresponding targets. During FL, only the shallow transformer layers are trained, keeping resource consumption modest.
In our experiments, a small transformer with two encoder layers, two decoder layers, and a hidden dimension of 32 achieves a prediction mean squared error below 0.5.

\begin{figure}
    \centering
    \vspace{1mm}
    \includegraphics[width=0.6\linewidth]{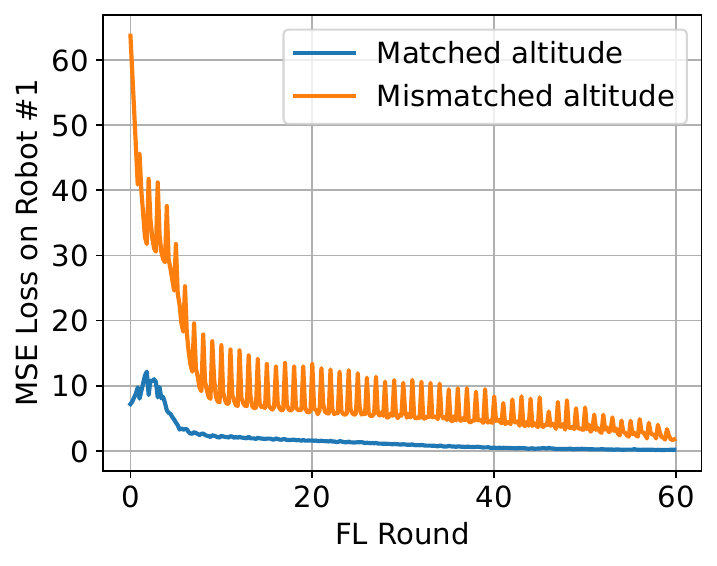}
    \vspace{-3mm}
    \caption{The MSE loss on robot \#1 in the motivation study. The two curves correspond to the cases where robot \#2 flies at the same altitude or at a different altitude from robot \#1.}
    \label{fig:height_motivation}
    \vspace{-5mm}
\end{figure}

The second challenge is data heterogeneity across distributed robots. In multi-robot target tracking, heterogeneity can arise from differences in target assignments and the fields of view, which is largely affected by drone positions including both horizontal location and altitude. Since target assignments are fixed, positional differences become the main source of heterogeneity.
To study its impact in FL, we experiment with two robots monitoring the same target at different horizontal locations,
flying at the matched and mismatched altitudes.
Fig.~\ref{fig:height_motivation} shows the MSE loss of target trajectory prediction on robot \#1 for both cases.
When the robots fly at different altitudes, the MSE loss on one robot exhibits fluctuations. These fluctuations arise from discrepancies between locally trained and globally aggregated models, highlighting data heterogeneity. The results indicate that differences in flight altitudes have a stronger impact on FL than horizontal position. To address this, \Method incorporates an altitude-invariant design to mitigate data heterogeneity.

Inspired by dynamics prediction with meta-learning~\cite{wang2022meta}, we incorporate adaptive instance normalization (AdaIN) to adapt the transformer $\theta$ using flight altitudes $h^i_t$. AdaIN, originally applied in neural style transfer~\cite{karras2019style}, adjusts normalization layers based on auxiliary information, in our case the flight altitude.
Let $z$ be the input to a normalization layer, with channel $z_k$ having mean $\mu(z_k)$ and standard deviation $\sigma(z_k)$. For each AdaIN layer, we compute a style vector $s = (\mu_k, \sigma_k)_k = A h^i_t + b$, where $A$ and $b$ are learnable parameters. The output is then computed as:
\begin{equation}
o_k = \sigma_k \frac{z_k - \mu(z_k)}{\sigma(z_k)} + \mu_k. \label{eq:adain}
\end{equation}
This operation shifts the normalized outputs according to the learned projection of flight altitude, making the model altitude-invariant.
We also recalibrate horizontal positions by expressing them relative to the $(x,y)$ location at time $t$, thus removing the dependence on absolute coordinates. 
Collectively, these design choices enforce position invariance in the learned model and accelerate FL convergence. 

\subsection{Trajectory Planning for UAVs}
\label{sec:method-control}

Given the predicted target trajectories, the next step is to compute actions that enable the UAV team to effectively trace and monitor the targets, as illustrated in Fig.~\ref{fig:method}. 
Unlike previous works based on measurement models of sensors~\cite{li2024resilient,ramachandran2020resilience}, \Method instead employs a neural network to predict the future target trajectories.
As the neural network model is deep and strongly non-convex, it is extremely challenging to derive the measurement model and estimate the uncertainty of each prediction.
To achieve robust tracking under these conditions, trajectory planning needs to therefore jointly (1) estimate and propagate the uncertainty associated with the learned predictions and (2) generate optimal control inputs which balance prediction quality and tracking performance.

We estimate the prediction uncertainty using an adaptive innovation-based Extended Kalman Filter. 
The predicted trajectory $\hat{\mathbf{z}}^i_{j,t_f}$ of target $j$ from robot $i$ is used as a noisy measurement of the true state.
The innovation, i.e., the difference between the prediction and the network output, is used to update the measurement noise covariance $R^i_j$ online via an exponential moving average. A Mahalanobis distance~\cite{mahalanobis2018generalized}–based gating step filters out outliers, keeping only statistically consistent innovations. The adapted $R^i_j$ is then fed back into the Kalman update, enabling the filter to adjust trust in each robot’s prediction and achieve robust multi-robot fusion under varying measurement quality.

We define a custom cost function $Q$ to balance tracking accuracy and future perception performance. Suppose $\mathbf{p}_{t_f}$ denotes the positions of all robots at time $t$, and $\hat{\mathbf{z}}_t$ denotes the predicted target trajectories from the adaptive Kalman Filter. The new cost function $Q$ is defined as:
\begin{equation}
Q(t+1) = \max \left (\|\mathbf{p}_{t+1} - \hat{\mathbf{z}}_{t+1}\|, h \right ) + \alpha \mathrm{Tr} (P_{t+1}), \label{eq:cost_func}
\end{equation}
where $P_{t+1}$ is the state covariance matrix after the prediction step, $\alpha$ is a weighting factor, $h$ is a minimum distance threshold. The first term reduces uncertainty by penalizing the trace of the covariance. The second term keeps robots close to targets, improving prediction accuracy and future perception quality from the drone-mounted cameras. The threshold $h$ prevents robots from getting too close and only observing partial targets. With this cost function, control inputs are obtained by solving Equation~\eqref{eq:control}.

\begin{figure*}[t]
    \centering
    \vspace{1mm}
    \includegraphics[width=0.74\linewidth]{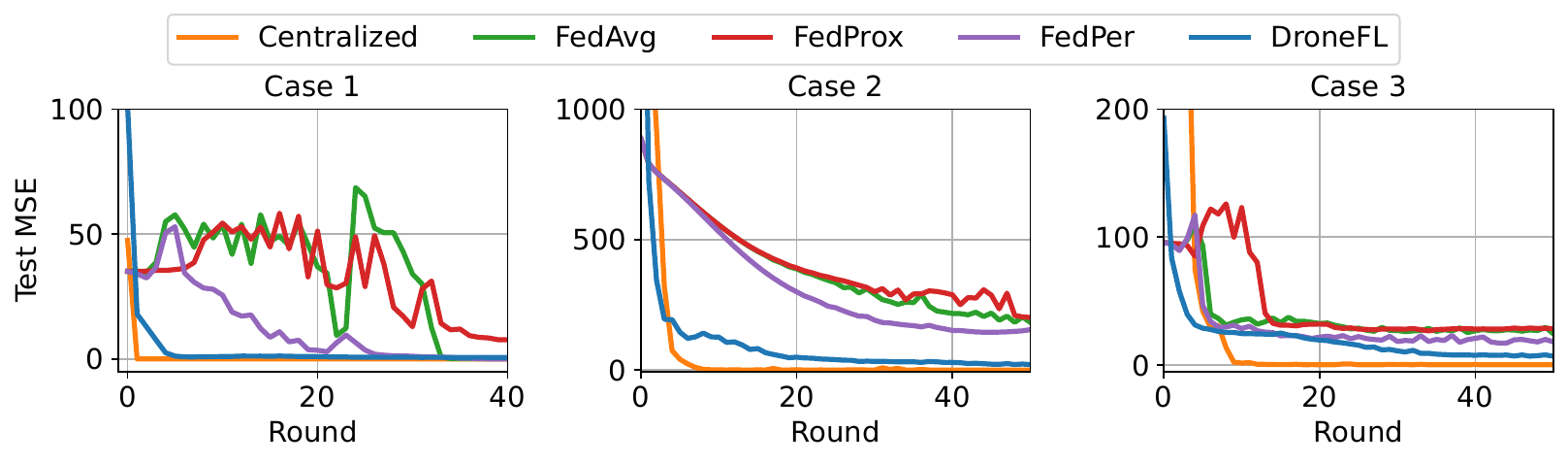}
    \vspace{-4mm}
    \caption{The average MSE loss across all UAVs in Case 1 (two robots, two targets, constant speed), Case 2 (three robots, three targets, random walk) and Case 3 (four robots, six targets, constant speed). Cases 2 and 3 are more challenging than Case 1 as reflected by higher MSE.} 
    \label{fig:fl_result}
    \vspace{-4mm}
\end{figure*}
\section{Evaluation}
\label{sec:evaluation}

\subsection{Experimental Setup}

\textbf{Simulation setup:} We focus on agriculture as a case study, while \Method can be easily extended to other scenarios and target types. We developed a simulated agricultural farm of size 160$\times$120 m in Gazebo~\cite{koenig2004design}. 
Each UAV flies at a distinct fixed altitude to avoid conflicts and is pre-assigned to monitor a specific subset of ground targets, ordered by altitude from highest to lowest.
The ground targets are modeled after agricultural machinery performing typical field operations in crop fields and orchards.
We evaluate \Method in several different environmental settings, as shown in Fig.~\ref{fig:teaser}. 
In a particular environment, we simulate the following cases:
\begin{enumerate}[label=(\arabic*),noitemsep,nolistsep]
\item \textit{Case 1:} Two robots monitor two targets moving at a constant speed of 0.5 m/s. Robot-target assignments are: $\mathcal{N}_1=\{1\}$, $\mathcal{N}_2=\{2\}$. This case provides the simplest feasibility check.
\item \textit{Case 2:} Three robots monitor three targets moving at a constant speed of 0.5 m/s with added random walk (each axis sampled from $\mathcal{N}(0,1)$). Robot-target assignments are: $\mathcal{N}_1=\{1,2,3\}$, $\mathcal{N}_2=\{1,2\}$, and $\mathcal{N}_3=\{2,3\}$. This setup simulates practical disturbances such as uneven terrain or obstacles.
\item \textit{Case 3:} A setup with four robots monitoring six targets moving at 0.5 m/s. Assignments are: $\mathcal{N}_1=\{1,2,3,$ $4,5,6\}$, $\mathcal{N}_2=\{1,2\}$, $\mathcal{N}_3=\{3,4\}$, and $\mathcal{N}_4=\{5,6\}$.
\end{enumerate}
The robot–target assignments ensure that each target is observed by at least one UAV, and sometimes two, providing redundancy for more reliable tracking.
Input images are resized to $640\times640$ pixels. Odometry is a 12-dimensional vector. Predicted target positions are 2D. All data are synchronized and recorded at 5 Hz. The data rate can be adjusted in the future based on target speed and application requirements.
All simulations are conducted on a Linux desktop equipped with an Intel Core Ultra 9 CPU (3.7-5.7 GHz), 64 GB RAM, and an NVIDIA GeForce RTX 5090 GPU.

\textbf{Framework setup:}
\Method is implemented in PyTorch.
For trajectory prediction, we employ a hybrid architecture that integrates a pretrained YOLOv11n detector~\cite{yolo-rpi5} with a transformer comprising two encoder layers and two decoder layers, with hidden dimension 16, feedforward dimension 32, and dropout rate $0.1$. 
Each UAV performs 10 epochs of local training using Adam~\cite{kingma2017adam} at a learning rate of $1e^{-3}$. During aggregation, updated models from all robots are combined. 
The weights in the cost function are set to $\omega=0.005, \alpha=50$.
FL is triggered every 100 seconds in simulation.
Code will be made publicly available upon publication.

\textbf{Metrics:}
We evaluate FL performance using the mean squared error (MSE) between predicted and ground-truth target positions. For end-to-end tracking, we evaluate the average trajectory prediction error and tracking distance for all UAV–target pairs. We also evaluate tracking uncertainty, control effort reflected in speed changes, and the total cost defined in Equation~\eqref{eq:obj}. We conduct hardware experiments by deploying \Method on a Raspberry Pi 5~\cite{rpi5},
a representative computing unit for UAVs, to measure the computational latency, energy consumption, and communication cost between each UAV and the cloud. The energy consumption is measured by the Hioki 3334 powermeter~\cite{powermeter}. 

\textbf{Baselines:}
We compare \Method with efficient FL baselines: \textbf{(i) FedAvg~\cite{mcmahan2017communication}}, the canonical FL method; \textbf{(ii) FedProx~\cite{li2020fedprox}}, which mitigates data heterogeneity via a regularization term in local training; and \textbf{(iii) FedPer~\cite{arivazhagan2019federated}}, which addresses heterogeneity by personalizing the final layer.
For fair comparison, all FL baselines use YOLO bounding boxes as input, the same as \Method.
For end-to-end tracking, since \Method is the first FL framework for multi-UAV target tracking, we compare it against two ablation setups that are commonly used in multi-UAVs: \textbf{(iv) centralized training}, where all robots send raw images and odometry to the cloud to train a large transformer model for future target prediction; and \textbf{(v) fixed distributed perception}, where local models are pretrained and remain unchanged during evaluation. In both cases, the predicted trajectories are processed by the same Extended Kalman Filter and controller as \Method, with necessary adaptations.

\begin{table*}[t]
\footnotesize 
\centering
\vspace{2mm}
\caption{\footnotesize Comparison of end-to-end tracking performance across different cases with \Method and the ablation setups. Bolded text indicates the best end-to-end tracking performance indicated by the prediction error and tracking distance.}
\vspace{-2mm}
\begin{tabular}{cc|c|cc|ccc}
\toprule
\textbf{Case} & \textbf{Setup} & \textbf{Method} & \textbf{Pred. Error} ($\downarrow$) & \textbf{Track. Distance} ($\downarrow$) &
\textbf{Uncertainty} ($\downarrow$) &
\textbf{Speed Change} ($\downarrow$) &  \textbf{Total Cost} ($\downarrow$) \\
& &  & $[m^2]$  & $[m^2]$  & $[m/s]$ & $[m^2 + (m/s)^2]$   &   \\
\midrule
\multirow{3}*{1} & 2 robots & Centralized & $\textbf{0.99} \pm \textbf{0.40}$ & $\textbf{14.82} \pm \textbf{0.35}$ & $0.41 \pm 2.48$ & $0.00 \pm 0.00$ & $22.37 \pm 0.30$ \\
& 2 targets & Fixed Distributed & $2.78 \pm 1.07$ & $24.65 \pm 0.40$ & $0.31 \pm 1.32$ & $0.40 \pm 0.86$ & $31.71 \pm 1.38$ \\
& Constant speed & \Method & $2.59 \pm 0.84$ & $24.54 \pm 0.37$ & $0.29 \pm 1.32$ & $0.46 \pm 1.04$ & $31.12 \pm 0.88$ \\
\hline
\multirow{3}*{2} & 3 robots & Centralized & $24.52 \pm 2.10$ & $62.99 \pm 3.71$ & $0.27 \pm 1.68$ & $0.33 \pm 0.66$ & $47.77 \pm 0.54$ \\
 & 3 targets & Fixed Distributed & $12.94 \pm 4.80$ & $54.10 \pm 2.26$ & $0.35 \pm 1.37$ & $0.51 \pm 1.02$ & $72.48 \pm 5.80$ \\
 & Random walk & \Method & $\textbf{2.17} \pm \textbf{1.19}$ & $\textbf{52.60} \pm \textbf{0.32}$ & $0.31 \pm 1.27$ & $0.63 \pm 0.57$ & $59.54 \pm 1.46$ \\
\hline
\multirow{3}*{3} & 4 robots & Centralized & $67.94 \pm 3.34$ & $142.12 \pm 5.48$ & $0.20 \pm 1.02$ & $0.06 \pm 0.30$ & $61.71 \pm 0.56$ \\
& 6 targets & Fixed Distributed & $14.50 \pm 3.87$ & $58.44 \pm 0.81$ & $12.48 \pm 58.77$ & $0.62 \pm 1.23$ & $78.32 \pm 11.07$ \\
 & Constant speed & \Method & $\textbf{4.14} \pm \textbf{2.21}$ & $\textbf{55.76} \pm \textbf{0.65}$ & $12.02 \pm 57.50$ & $0.54 \pm 1.20$ & $72.13 \pm 4.70$ \\
\bottomrule
\vspace{-6mm}
\label{tbl:end2end}
\end{tabular}
\end{table*}

\subsection{Offline FL Performance}
\label{sec:fl-results}
We first evaluate offline trajectory prediction performance using pre-collected data and randomly initialized models.
This experiment uses the simulated crop field environment shown in Fig.~\ref{fig:teaser}b.
Data is collected by manually controlling robots to monitor their assigned targets.
We perform 80/20 train-test split on each robot's data.
For \Method and all FL baselines, we set the global rounds to 50, an input window of $L_p=10$, and an output window of $L_f=5$, meaning the model predicts a 1-second trajectory from the past 2 seconds.

Fig.~\ref{fig:fl_result} shows the average test MSE across all UAVs.
All FL baselines (FedAvg, FedProx, FedPer) struggle to converge, though FedPer achieves a lower MSE by personalizing the last layer.
These results underscore the difficulty of learning a uniform model under heterogeneous data.
\Method reduces the MSE by 47\%-96\% compared to the FL baselines. 
These gains are even more significant in the challenging Cases 2 and 3, which involve more heterogeneous data due to a larger swarm of UAVs and irregular target dynamics.
These improvements demonstrate the effectiveness of the position-invariant design in aligning model parameters and mitigating heterogeneity.
\Method also approaches the performance of centralized training, an upper bound with access to all raw images and odometry from every UAV. However, as discussed in Sec.~\ref{sec:efficiency-results}, centralized training requires excessive energy and incurs impractical communication costs.

\subsection{End-to-End System Performance}
\label{sec:system-results}

We perform end-to-end tracking evaluation to answer the question: what benefits can federated learning bring to multi-robot target tracking? \Method is compared to centralized training and fixed distributed models. All methods use models pretrained in the crop field environment (Fig.\ref{fig:teaser}b) and deployed in the orchard environment (Fig.\ref{fig:teaser}c), aligning with the practical challenge of mismatched training data. Each method uses an input window of $L_p=10$ and an output window of $L_f=1$. In \Method, the fused target trajectories are used for local training, one FL round is triggered every 100 seconds in simulation.
Table~\ref{tbl:end2end} presents the end-to-end tracking performance of all methods across three cases. Trajectory prediction error and tracking distance are key tracking performance indicators, with the best values highlighted in bold in the table. Good end-to-end tracking performance requires a tight coupling between trajectory prediction and UAV planning. Uncertainty, speed change, and total cost assess the stability of the planning process.

\textbf{\Method vs. Centralized:} Centralized training performs well in the simplest Case 1 but fails to scale to more complex cases. In Cases 2 and 3, although the centralized baseline achieves minimal speed changes, tracking uncertainty, and total cost, it completely loses track of the targets, as reflected by the extremely large prediction errors and tracking distances. This failure stems from the poor generalizability of visual-based end-to-end training:
when the visual background or control strategies change, predictions drift, and the controller reacts to these drifted predictions, ultimately losing the targets.
In contrast, \Method scales and generalizes effectively in challenging cases. Its disaggregated design separates the YOLO visual model from the trajectory prediction model. YOLO handles visual processing while trajectory prediction uses only the detected bounding boxes. This structure allows \Method to adapt to different agricultural environments and larger UAV swarms.

\textbf{\Method vs. Fixed Distributed:} Compared to distributed trajectory prediction with fixed models, \Method reduces prediction error by 6\%-83\% and tracking distance by 0.4\%-4.6\%, demonstrating the advantage of on-device FL. FL allows models to adapt quickly to new environments and changing target dynamics.
On the planning side, the cost function balances tracking performance (by minimizing uncertainty and distance) and control stability (by minimizing speed changes). In the simulation, we observe that adding the speed-change term in the cost function is critical for stable flight, which in turn supports accurate trajectory prediction.
Overall, the closed-loop design in \Method provides continuous monitoring and more accurate knowledge of all targets compared to using distributed fixed models on the UAVs.

\begin{table}[t]
\centering
\setlength{\tabcolsep}{3.5pt}
\caption{Comparison of methods in efficiency.  Bolded text indicates better efficiency results.}
\vspace{-2mm}
\begin{tabular}{lcc}
\toprule
Metrics & Centralized & \Method \\
\cmidrule(lr){2-2} \cmidrule(lr){3-3}
 & RTX 5090 GPU & RPi 5 \\
 \midrule
Communication cost (KB/sec) & $2.46\times10^3$ & \textbf{1.56} \\
Infer. latency per sample (sec) & 0.63 & \textbf{0.13} (Y.) + \textbf{0.01} (T.)$^1$ \\
Infer. energy per sample (J) & 38.85 &  \textbf{0.78 }(Y.) + \textbf{0.11} (T.)$^1$  \\
Train. latency per epoch (sec) & 33.61 & \textbf{1.76} \\
Train. energy per epoch (J) & 4499.03 & \textbf{14.11 }\\
\bottomrule
\multicolumn{3}{l}{$^1$``Y.'' denotes the YOLO model and ``T.'' denotes the transformer model} \\
\vspace{-8mm}
\label{tbl:efficiency}
\end{tabular}
\end{table}

\subsection{Communication Costs and Computational Efficiency}
\label{sec:efficiency-results}
Table~\ref{tbl:efficiency} compares \Method with centralized training in terms of efficiency. Centralized training and inference are executed on a desktop with an RTX 5090 GPU (cloud setting), while \Method runs both on a Raspberry Pi 5~\cite{rpi5}, representative of UAV hardware.
For communication, the centralized baseline requires an impractical costs of 2.46 MB/sec to transmit raw images from UAVs to cloud, while communication in \Method is minimal: only predicted target positions are sent in real time, and the transformer model is updated every 100 seconds, reducing the cost to just 1.56 KB/s.
For computation, inference must keep pace with 5 Hz data collection, while training occurs every 100 seconds. The centralized baseline fails to meet this real-time requirement even on the cloud, with 0.63 s inference latency, and its training is energy-intensive due to the large transformer. By contrast, \Method uses a frozen YOLO backbone and a lightweight transformer, achieving real-time inference (0.14 s per sample) on a Raspberry Pi 5 and completing one local training epoch in 1.76 s. These results highlight \Method’s efficiency and feasibility on resource-constrained UAVs.

\section{Conclusion}
\label{sec:conclusion}

While federated learning has the potential to improve learning effectiveness in multi-UAV target tracking, progress in this area is hindered by challenges in limited hardware resources, heterogeneous data distribution across UAVs and achieving robust coupling between target trajectory prediction and UAV trajectory planning. 
To this end, we propose \Method, the first FL framework for multi-UAV visual target tracking.
\Method uses a lightweight architecture with a frozen YOLO model and shallow transformer, and addresses FL data heterogeneity via altitude-based adaptive instance normalization. For target tracking, all UAV predictions are fused in the cloud to generate future trajectories using a custom cost function that balances tracking accuracy and stability.
Comprehensive simulations show that \Method reduces target prediction error and tracking distance over non-FL distributed methods. \Method also achieves real-time prediction on a Raspberry Pi 5 while requiring only lightweight communication with the cloud.

\bibliographystyle{IEEEtran}
\bibliography{references}

\end{document}